\documentclass[letterpaper, 10 pt, conference]{ieeeconf}  % Comment this line out if you need a4paper

\IEEEoverridecommandlockouts                              % This command is only needed if 
\usepackage{cite}
\usepackage{pdflscape}
\usepackage{adjustbox}

\usepackage{algorithm}

\usepackage[
]{algpseudocode}

\usepackage{afterpage}

\usepackage{graphicx}

\usepackage[
    labelformat=simple
]{subcaption}

\usepackage{amsmath}

\usepackage{amssymb}

\usepackage{bbm}

\usepackage{mathrsfs}

\usepackage{mathtools}

\usepackage{xfrac}

\usepackage[Gray,squaren,thinqspace,thinspace]{SIunits}
\usepackage{array}

\usepackage{booktabs}

\usepackage{multirow}

\usepackage{tabu}

\usepackage{tabularx}

\usepackage{tabulary}

\usepackage{pifont}

\usepackage{lipsum}
\usepackage{xcolor}
\usepackage{microtype}
\usepackage{hyperref}
\usepackage[normalem]{ulem}

\usepackage{changepage}
\usepackage{datetime}
\usepackage{tabularx}
\usepackage{xcolor}

\newenvironment{base_cover_letter}[3]{
    \begin{titlepage}
    \def\paperTitle{#1}
    \def\journalName{#2}
    \def\authorName{#3}
    \setlength{\parindent}{0pt}
    \setlength{\parskip}{1.5em}
    \fontsize{10pt}{12pt}\selectfont
    \begin{center}
        \LARGE \textbf{Cover Letter}
    \end{center}
    \bigskip
}{
    Sincerely, \\
    \authorName{}
    \end{titlepage}
}

\AtBeginDocument{\colorlet{defaultcolor}{.}}

\title{\LARGE \bf
Precise Pick-and-Place using Score-Based Diffusion Networks
}

\author{Shih-Wei Guo$^{1}$, Tsu-Ching Hsiao$^{12}$, Yu-Lun Liu$^{3}$, and Chun-Yi Lee$^{12}$% <-this % stops a space
\thanks{$^{1}$Elsa Lab, National Tsing Hua University, Hsinchu City, Taiwan.}%
\thanks{$^{2}$Elsa Lab, National Taiwan University, Taipei City, Taiwan.}%
\thanks{$^{3}$National Yang Ming Chiao Tung University, Hsinchu City, Taiwan.}%
}

\usepackage{eso-pic}

\newcommand{\copyrighttext}{%
\footnotesize \textcopyright\ 2024 IEEE. Personal use of this material is permitted. Permission from IEEE must be obtained for all other uses, in any current or future media, including reprinting/republishing this material for advertising or promotional purposes, creating new collective works, for resale or redistribution to servers or lists, or reuse of any copyrighted component of this work in other works.}

\newcommand{\copyrightnotice}{%
  \AddToShipoutPicture*{%
    \AtTextUpperLeft{%
      \raisebox{0cm}[0pt][0pt]{%
        \begin{minipage}{\textwidth}
          \centering \color{red}\fbox{\parbox{\dimexpr\textwidth-\fboxsep-\fboxrule\relax}{\copyrighttext}}
        \end{minipage}}}}}

\begin{document}

\maketitle
\thispagestyle{empty}
\pagestyle{empty}
\copyrightnotice

\newcommand{\Log}{\text{Log}}
\newcommand{\Exp}{\text{Exp}}

%%%%%%%%%%%%%%%%%%%%%%%%%%%%%%%%%%%%%%%%%%%%%%%%%%%%%%%%%%%%%%%%%%%%%%%%%%%%%%%%
\begin{abstract}
In this paper, we propose a novel coarse-to-fine continuous pose diffusion method to enhance the precision of pick-and-place operations within robotic manipulation tasks. Leveraging the capabilities of diffusion networks, we facilitate the accurate perception of object poses. This accurate perception enhances both pick-and-place success rates and overall manipulation precision. Our methodology utilizes a top-down RGB image projected from an RGB-D camera and adopts a coarse-to-fine architecture. This architecture enables efficient learning of coarse and fine models. A distinguishing feature of our approach is its focus on continuous pose estimation, which enables more precise object manipulation, particularly concerning rotational angles. In addition, we employ pose and color augmentation techniques to enable effective training with limited data. Through extensive experiments in simulated and real-world scenarios, as well as an ablation study, we comprehensively evaluate our proposed methodology. Taken together, the findings validate its effectiveness in achieving high-precision pick-and-place tasks.
\end{abstract}

%%%%%%%%%%%%%%%%%%%%%%%%%%%%%%%%%%%%%%%%%%%%%%%%%%%%%%%%%%%%%%%%%%%%%%%%%%%%%%%%
\section{Introduction}
\label{sec:introduction}
Pick-and-place tasks, which involve picking and placing objects with accuracy in terms of both position and orientation, are crucial in various industrial applications such as assembly lines, handling electronic components, and transporting semiconductor wafers. Failure in these tasks can lead to severe consequences including production line halts and financial losses. Common industry practices to enhance object recognition and position tracking for pick-and-place tasks involve using 2D barcodes, such as DataMatrix, ArUco~\cite{romero2018speeded}, and AprilTag~\cite{olson2011apriltag}. Nevertheless, these approaches often necessitate modifications to the environment or objects, which limits their versatility. Some prior approaches rely on object models~\cite{narayanan2016discriminatively, kehl2017real, gualtieri2021robotic}, however, accurate or customized models might not always be accessible. Recent advancements in deep learning have led to the emergence of end-to-end models~\cite{levine2016end, rahmatizadeh2018vision, kalashnikov2021mt, berscheid2020self}, which directly translate images into actions. These models address the above challenges and expand the possibilities in pick-and-place tasks. However, such models sometimes require substantial amounts of data for training.

Training pick-and-place models in an end-to-end manner typically demands extensive data, which can be challenging to acquire, particularly in real-world scenarios. The process of gathering data in real-world settings could be resource-intensive and time-consuming~\cite{kalashnikov2021mt}. While some methods employ self-supervised deep reinforcement learning (DRL) to achieve this objective, they lack the ability to precisely place objects~\cite{zeng2018learning}. Moreover, due to the insufficiency of real-world data, certain experiments are confined to simulated environments. This confinement leaves their real-world performance uncertain~\cite{huang2024fourier}. While approaches such as Transporter Network~\cite{zeng2021transporter} have shown promise in using less number of demonstrations in simulation, their application in the real world still relies on extensive manual data collection. Moreover, Transporter Network's discrete rotational outputs with a limited resolution constrain its effectiveness in scenarios requiring continuous rotational outputs and precise manipulation. Building upon Transporter Network, some methods, such as those leveraging equivariant network~\cite{fu2023multi, huang2024leveraging, huang2024fourier}, aim to enhance training efficiency and reduce data requirements. However, challenges persist regarding discrete rotational outputs and limited angular resolution. To address the discrete rotational resolution problem, the study in~\cite{soti2023train} introduced iterative angle refinement, albeit maintaining a discrete nature. The authors in~\cite{berscheid2020self} devised a self-supervised learning method to generate abundant data and achieve high precision. However, this method requires the integration of force sensors, which increases complexity. The work~\cite{simeonov2023shelving} presented diffusion networks for continuous pose outputs, but it relies on point cloud inputs. Despite their effectiveness, these methods necessitate additional sensor requirements, which can be restrictive in certain applications.

In light of these issues, we introduce a new coarse-to-fine continuous pose diffusion method designed to significantly enhance precision and success rates in pick-and-place tasks. Our methodology leverages diffusion networks that are capable of generating continuous pick-and-place poses. By utilizing RGB images as inputs, which are projected from a top-down perspective via an RGB-D camera, our methodology eliminates the necessity for additional sensors. Moreover, through the integration of pose augmentation techniques, our method demonstrates exceptional efficacy with a small amount of training data. The contributions are summarized as follows:
\begin{itemize}
\item We introduce a coarse-to-fine approach for generating continuous pick-and-place poses using diffusion networks.
\item We demonstrate the effectiveness of our approach by achieving high precision and success rates with a small amount of training data in both simulated and real-world environments. Our results surpass the performance of the baselines, which highlight the potential of our method.
\item We require only top-down projected RGB images, offering a cost-effective and accessible solution.
\end{itemize}

\section{Related Work}
\label{sec:related_work}
\subsection{Pick-and-Place}
In the realm of perception for manipulation, object detection and object pose estimation are widely used for determining the position of target objects. Model-based approaches in object pose estimation, exemplified by~\cite{zhu2014single, narayanan2016discriminatively, kehl2017real}, offer precise estimations and are particularly suitable for pick-and-place tasks demanding high accuracy, as highlighted in~\cite{gualtieri2021robotic}. However, they either rely on 3D object models or require point cloud data, which limits their applicability in practical scenarios where such resources are unavailable.
\subsection{Transporter Network and Its Successor}
In response to the demand for model-free pick-and-place capabilities, various approaches have emerged. The Transporter Network~\cite{zeng2021transporter}, for example, introduced an end-to-end method for pick-and-place tasks using minimal demonstrations. It leverages three fully convolutional networks: one for predicting the pick position and the other two for determining the place position and rotation. Transporter Network has been widely adopted in other works. The study in~\cite{seita2021learning} incorporates image-based goal conditioning and is capable of handling deformable objects, CLIPort~\cite{shridhar2022cliport} integrates language conditioning to learn multi-task policy, while another work~\cite{lim2022multi} introduces sequence conditioning to solve multi-task long horizon problems, and the authors in~\cite{soti2023train} employ iterative inference methods to enhance angular resolution. To enhance sample efficiency, methods such as~\cite{fu2023multi, huang2024leveraging, huang2024fourier} adopt equivariant models to exploit the symmetry inherent in pick-and-place tasks. Unfortunately, these approaches produce discrete pick-and-place poses. In contrast, our method aims to estimate continuous poses, which offers a distinct advantage.
\subsection{Diffusion Models and Its Application in Manipulation}
Recent advancements in diffusion generative models~\cite{ho2020denoising, song2021denoising} offer a promising avenue for learning pick-and-place distributions and generating continuous pose outputs. These models excel in capturing complex data distributions~\cite{song2021scorebased}, making them well-suited for handling multimodal distributions~\cite{song2019generative}. Moreover, the iterative sampling process inherent in diffusion models endows them with robust tolerance to data noise. This robustness makes them suitable for real-world applications across various domains, especially for robotic manipulations~\cite{simeonov2023shelving, mishra2023reorientdiff, xian2023chaineddiffuser, chen2023playfusion}. Diffusion-EDFs~\cite{ryu2024diffusion} introduces an equivariant diffusion model on $SE(3)$ to enhance data efficiency, but it requires additional collection of grasp object's point clouds. The study in~\cite{ha2023scaling} uses a Large Language Model (LLM) and diffusion policy~\cite{chi2023diffusionpolicy} to generate manipulation trajectories. However, most of these robotic manipulation approaches rely on point clouds. In contrast, our method stands out by relying only on a top-down RGB image projected from an RGB-D camera. This exclusive reliance on RGB images simplifies data acquisition and therefore improves practical applicability.
\section{Background}
\label{sec:background}
\subsection{Score-Based Generative Models}
Score-based generative models (SGMs)~\cite{song2021scorebased} provide a practical framework for recovering an underlying data distribution $p_\text{data}(x)$ from independent and identically distributed (i.i.d.) samples $\{x_n | x_n\sim p_\text{data}\}^N_{n=1}$. In Noise Conditional Score Network (NCSN)~\cite{song2019generative}, the data distribution is gradually transformed into a tractable prior distribution, typically a Gaussian distribution $\mathcal{N}(x;0,\sigma^2_L)$, by the \textit{forward} process. The \textit{forward} process is an iterative process that adds a set of noises $\{\sigma_i\}^L_{i=1}$ to the samples, where $L$ is the total number of diffusion steps, with corresponding perturbation kernels $p_{\sigma_i}(\tilde{x}|x)=\mathcal{N}(\tilde{x};x, {\sigma_i}^2 )$, where $\sigma_1 < \sigma_2 < \cdots < \sigma_L$. A score network $s_\theta(x; \sigma)$ parameterized by $\theta$ is trained to estimate the (Stein) \textit{score}~\cite{liu2016kernelized} of the perturbation kernel, represented as the gradient of its logarithm $\nabla_{\tilde{x}}\log p_\sigma(\tilde{x}|x)$, via a Denoising Score Matching (DSM)~\cite{vincent2011connection} objective. During the generation stage, NCSN utilizes Langevin Markov Chain Monte Carlo (MCMC) method as the reverse process to iteratively generate samples from the prior distribution.

\subsection{Score-Based Pose Diffusion Models}
Based on the above, the author in~\cite{hsiao2024confronting} extends the concept to operate on the Lie groups and the rotational space, specifically $SO(3)$ and $SE(3)$~\cite{Deray-20-JOSS}. This shows superior accuracy and effectiveness in resolving pose ambiguity encountered in 6D object pose estimation. Assuming a Lie group $\mathcal{G}$ with its associated Lie algebra $\mathfrak{g}$, and considering group elements $X, \tilde{X}\in\mathcal{G}$, the transition between these elements is defined as $\tilde{X}=X\Exp(z)$,\footnote{Exponential map $\Exp: \mathfrak{g}\to\mathcal{G}$; logarithm map $\Log: \mathcal{G}\to\mathfrak{g}$; composition $\circ: \mathcal{G}\times\mathcal{G}\to\mathcal{G}$, in shorthand: $X\circ Y=XY$.} where $z\in\mathfrak{g}$ and $z\sim \mathcal{N}(0, \sigma^2 I)$. This instantiates a perturbation kernel expressed as the following:
\begin{equation}
{\scriptsize
    \begin{split}
    p_\Sigma(\tilde{X}|X) &:= \mathcal{N}_\mathcal{G}(\tilde{X};X, \Sigma)\\
    &\triangleq \frac{1}{\zeta(\Sigma)}\exp\left( -\frac{1}{2}\Log(X^{-1}\tilde{X})\Sigma^{-1}\Log(X^{-1}\tilde{X}) \right),
    \end{split}
}
\label{eq:perturbation-kernel}
\end{equation}
where $\Sigma$ represents the covariance matrix with diagonal entries denoted by $\sigma$ to indicate the scale of perturbation. The normalization constant $\zeta(\Sigma)$ ensures proper scaling. The \textit{score} of Eq.~(\ref{eq:perturbation-kernel}) with respect to $\tilde{X}$ is defined as follows:
\begin{equation}
    \nabla_{\tilde{X}} \log p_\sigma(\tilde{X}|X) = -\frac{1}{\sigma^2}\mathbf{J}_r^{-\top}(z)z,
\end{equation}
where $\mathbf{J}_r^{-\top}$ is the inverse transpose of the right-Jacobian on $\mathcal{G}$. The author in~\cite{hsiao2024confronting} proves that the \textit{score} on $SE(3)$ can be represented through a closed-form approximation, defined as:
\begin{equation}
    s_X(\tilde{X}; \sigma)\triangleq -\frac{1}{\sigma^2}z,
    \label{eq:lie-score}
\end{equation}
where this approximation is termed as the surrogate Stein \textit{score}. Following this, $s_\theta(\tilde{X}; \sigma)$ is trained using the Lie group variant of the DSM objective $\mathcal{L}_\text{DSM}(\theta; \sigma)$, defined as follows:
\begin{equation}
{\scriptsize
    \begin{split}
    \mathcal{L}_\text{DSM}(\theta; \sigma) \triangleq \frac{1}{2} \mathbb{E}_{p_{\text{data}}(X)}\mathbb{E}_{\tilde{X} \sim \mathcal{N}_{\mathcal{G}}(X, \Sigma)} \left[ \left\| s_{\boldsymbol{\theta}}(\tilde{X}; \sigma) - s_X(\tilde{X}; \sigma) \right\|^{2}_{2} \right].
    \end{split}
}
\label{eq:lie-dsm}
\end{equation}
To draw samples from $s_\theta(\tilde{X}; \sigma)$, the reverse process is enacted through the Geodesic Random Walk~\cite{jorgensen1975central} on $\mathcal{G}$, defined as:
\begin{equation}
{\footnotesize
\begin{split}
    \tilde{X}^{(i-1)} = \tilde{X}^{(i)}\Exp(\epsilon_i s_\theta(\tilde{X}^{(i)}; \sigma_i) + \sqrt{2\epsilon_i} z^{(i)}), \quad z^{(i)}\sim \mathcal{N}(0, I).
\end{split}
}
\label{eq:random-walk}
\end{equation}

\section{Methodology}
\label{sec:methodology}
\subsection{Problem Statement}
The pick-and-place task involves estimating the position and orientation of both the object to be picked and the target location for placement, and subsequently utilizing a robotic arm for transporting the object to the designated target location and orientation. Given an RGB observation $\mathcal{I}\in \mathbb{R}^{H\times W\times 3}$ in a top-down view, we define the pick and place poses as random variables of a joint probability distribution $p(X, Y | \mathcal{I})$ conditioned on $\mathcal{I}$, where $(X, Y)\in SE(2)^2$ represents the pick and place poses residing in $SE(2)$ as we limit our working area to a top-down 2D space. In the following context, unless stated otherwise, we denote the pick pose as $X$ and the place pose as $Y$. Our objective is to recover the joint probability $p_D(X, Y|\mathcal{I})$ formed from a set of limited number of demonstrations $\{(X, Y, \mathcal{I})_k\}^{K}_{k=1}\subseteq D$ via score-based pose diffusion models, where $D$ is the demonstrations and $K$ is the number of demonstrations, and subsequently we use these models to estimate the pick-and-place poses from unseen observations. We then make use of a conventional motion-planning algorithm for controlling the robotic arm to execute pick-and-place operations based on the estimated poses. Following the approach outlined in~\cite{zeng2021transporter}, the observation is projected to a top-down view using the camera pose and the ground truth depth values. In contrast to prior methods that estimate from a limited set of predefined, discretized position and rotation values, our method possesses the capability to predict continuous pick-and-place poses.

\subsection{Framework Overview }
\begin{figure*}[t]
  \centering
  \includegraphics[width=\textwidth]{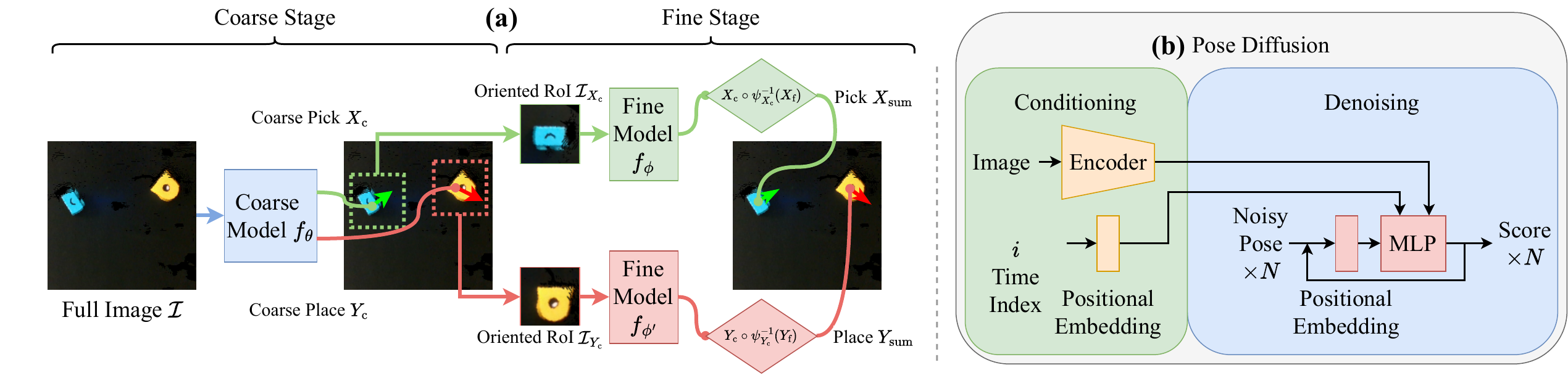}
  \caption{
  The proposed two-stage pose diffusion framework. (a) The coarse-to-fine stages for estimating the pick and place poses. (b) The pose diffusion models in (a) comprise a conditioning part and a denoising part.
  }
  \label{fig:architecture}
  \vspace{-1em}
\end{figure*}
Fig.~\ref{fig:architecture}~(a) illustrates an overview of the framework. Our framework takes a full image $\mathcal{I}$ of the workspace as input and predicts the corresponding pick and place poses. To ensure accuracy in pick-and-place operations, our framework utilizes a two-stage prediction approach comprising a \textit{coarse stage} and a \textit{fine stage}. In the coarse stage, our coarse model $f_\theta$ parameterized by $\theta$ concurrently estimates the coarse poses of the pick and place targets $(X_\text{c}, Y_\text{c})\in SE(2)^2$. We denote the procedure as $(X_\text{c}, Y_\text{c})\sim f_\theta(X_\text{c}, Y_\text{c}|\mathcal{I})$. Based on the predicted pick and place poses, the image $\mathcal{I}$ is transformed into the corresponding oriented region of interests (ORoIs) with affine transformation, denoted as $(\mathcal{I}_{X_\text{c}}, \mathcal{I}_{Y_\text{c}})\triangleq (\Psi_{X_\text{c}}(\mathcal{I}), \Psi_{Y_\text{c}}(\mathcal{I}))$, where $\Psi_Z: \mathcal{I}\to\mathcal{I}_{Z}$ indicates the affine transform of images giving an element $Z\in SE(2)$. In ORoIs, the targets are conceivably centered and facing a unified direction. We refer to the coordinate frame on ORoI as the ORoI space and use $\psi_Z: SE(2)\to SE(2)$ to indicate the mapping from the image coordinate frame to the ORoI space, with $\psi_Z^{-1}$ indicating its inverse mapping. In the fine stage, two fine models $f_\phi$ and $f_{\phi'}$ parameterized by $\phi$ and $\phi'$ respectively are utilized to estimate the refined poses, which represent the residuals between the pick and place poses and the centers of their corresponding ORoIs, based on the respective ORoI crops. Specifically, we denote the fine poses of pick and place targets as $(X_\text{f}, Y_\text{f})\in SE(2)^2$, and the processes are defined as $X_\text{f}\sim f_\phi(X_\text{f}|\mathcal{I}_{X_\text{c}})$ and $Y_\text{f}\sim f_{\phi'}(Y_\text{f}|\mathcal{I}_{Y_\text{c}})$. The final predictions of pick and place poses are subsequently estimated by aggregating the coarse and fine predictions using the equations expressed as follows:
\begin{equation}
X_\text{sum} = X_\text{c}\circ\psi_{X_\text{c}}^{-1}(X_\text{f}),\quad Y_\text{sum} = Y_\text{c}\circ\psi_{Y_\text{c}}^{-1}(Y_\text{f}).
\label{eq:pose-aggregation}
\end{equation}
The transport pose $\mathcal{T}$, defined as the transformation from a pick pose to a place pose, is then calculated as the composition of the two poses. The transport pose $\mathcal{T}$ is defined as follows:
\begin{equation}
\mathcal{T}=Y_\text{sum}\circ X^{-1}_\text{sum}.
\label{eq:transport}
\end{equation}
It is expected that the transformation of the full image into smaller ORoIs in the coarse stage allows for focused learning of the relevant pick-and-place regions and features in the fine stage. This transformation results in a substantial improvement in both learning efficiency and prediction accuracy. We elaborate on the benefits in Section~\ref{sec:results}.

\subsection{Extending Score-Based Pose Diffusion Models}
In our framework, we employ the modified version of score-based pose diffusion models~\cite{hsiao2024confronting} as our coarse model and fine models in the two stages for generating precise pick and place poses. As the pose diffusion models originally introduced by~\cite{hsiao2024confronting} operate on $SE(3)$, we reduce the dimensionality of the operating space to $SE(2)$ through a specific parametrization technique. This technique involves representing elements in $SE(3)$ as $(R, T)$ pairs, where $R=(\omega_x, \omega_y, \omega_z)\in SO(3)$ denotes the Euler angle representation of the rotational element in $SO(3)$, and $T=(\tau_x, \tau_y, \tau_z)\in \mathbb{R}^{3}$ represents translations along the $x$, $y$ and $z$ axes. Given our assumption of a top-down view in the workspace, we parametrize the $SE(2)$ space using a tuple $(\omega_z, \tau_x, \tau_y)$. Furthermore, we extend the operating space of the pose diffusion models from a single $SE(2)$ primitive space to a compositional space denoted as $SE(2)^N$, where $N$ is the number of composed $SE(2)$. Considering two compositional elements $(X_1, X_2, \cdots, X_N)\in SE(2)^N$ and $(\tilde{X}_1, \tilde{X}_2, \cdots, \tilde{X}_N)\in SE(2)^N$ with the relationship $\tilde{X}_n=X_n\text{Exp}(z_n), ~z_n\sim\mathcal{N}(0, \sigma_n I)$, we define the Gaussian perturbation kernel on $SE(2)^N$ using the following equation:
\begin{equation}
{\footnotesize
\begin{split}
    &p_\mathbf{\Sigma}\left(\tilde{X}_1, \tilde{X}_2, \cdots, \tilde{X}_N\middle|X_1, X_2, \cdots, X_N\right) \\
    &\qquad\qquad\triangleq \prod_{n=1}^N p_{\sigma_n}(\tilde{X}_n|X_n) =\prod_{n=1}^N \mathcal{N}(\tilde{X}_n; X_n, \sigma_n I),
\end{split}
}
\end{equation}
where $\mathbf{\Sigma}\in \mathbb{R}^{3N\times 3N}$ is the covariance matrix, $\sigma_n$ corresponds to the covariance for the distribution of the $n$-th primitive, and $\mathcal{N}(\tilde{X}_n;X_n, \sigma_n I)$ follows the definition in Eq.~(\ref{eq:perturbation-kernel}). An important assumption is the mutual independence of each element in the compositional set. This allows us to simplify the joint distribution to the multiplication of individual conditional distributions. Thus, the (Stein) \textit{score} with respect to $\tilde{X}_j$, $j \in \{1,\cdots,N\}$,  is reduced by following the property of logarithm as:
\begin{equation}
{\scriptsize
\begin{split}
    \nabla_{\tilde{X}_j}\log \left(\prod^N_{n=1}p_{\sigma_n}(\tilde{X}_n|X_n)\right) &= \nabla_{\tilde{X}_j}\left(\sum^N_{n=1}\log p_{\sigma_n}(\tilde{X}_n|X_n)\right)\\
    =\nabla_{\tilde{X}_j} \log p_{\sigma_j}(\tilde{X}_j|X_j)&=-\frac{1}{\sigma_j^2}\mathbf{J}^{-\top}_{r}(z_j)z_j.
\end{split}
}
\end{equation}
Following the definition of surrogate Stein \textit{score} in Eq.~(\ref{eq:lie-score}) and the DSM objective in Eq.~(\ref{eq:lie-dsm}), we define the DSM objective on $SE(2)^N$ as the summation of the individual DSM loss on each primitive space. We define the score network as $s_\theta(\tilde{X}_n; \sigma_n)$ and formulate the DSM objective as follows:
\begin{equation}
    \mathcal{L}_{\text{DSM-}N}(\theta; \sigma) = \sum^{N}_{n=1}\mathcal{L}^{(n)}_\text{DSM}(\theta; \sigma).
\label{eq:lie-dsm-n}
\end{equation}
Following the sampling procedure in~\cite{hsiao2024confronting}, we initially draw a noisy sample from a known prior distribution $\tilde{X}^{(L)}_{n}\in\mathcal{N}_{SE(2)}(0, \sigma_L I)$ in the corresponding $n$-th primitive space. We then execute the reverse process similar to Eq.~(\ref{eq:random-walk}) to iteratively denoise the sample across time steps $i=\{L, L-1,\cdots, 1\}$ using the estimated \textit{score} on individual primitive spaces. We formulate the reverse process on $SE(2)^N$ as:
\begin{equation}
{\footnotesize
\begin{split}
    \tilde{X}^{(i-1)}_n = \tilde{X}^{(i)}_n\Exp(\epsilon_i s_\theta(\tilde{X}^{(i)}_n; \sigma_i) + \sqrt{2\epsilon_i} z^{(i)}_n), \quad z^{(i)}_n\sim \mathcal{N}(0, I).
\end{split}
}
\label{eq:random-walk-n}
\end{equation}
In practice, the score network can be designed as a unified one, taking a composed element as input and generating multiple \textit{score} estimations corresponding to each primitive element.

\subsection{Architecture Design}
As discussed in the previous subsections, our framework models the pick and place poses as random variables of a joint distribution, defined as $(X, Y)\sim p(X, Y|\mathcal{I})$, conditioned on the image $\mathcal{I}$. Our objective is to recover the true distribution $p(X, Y|\mathcal{I})$ from its empirical counterpart $p_{D'}(X, Y|\mathcal{I})$, which is derived from a limited number of demonstrations $D'=\{(X, Y, \mathcal{I})_{k}\}^{K}_{k=1}\subseteq D$, with $K$ the number of demonstrations. To achieve this, we employ our extended score-based pose diffusion model on $SE(2)^N$. Fig.~\ref{fig:architecture}~(b) depicts the general architecture of our pose diffusion models. In the coarse stage, our coarse model $f_\theta(X_\text{c}, Y_\text{c}|\mathcal{I})$ is trained to fit the empirical distribution. This distribution is implicitly modeled using the inherent score network, defined as $s_\theta(\tilde{X}_\text{c}, \tilde{Y}_\text{c}|\mathcal{I}; \sigma)$, which operates on $SE(2)^2$. In the fine stage, we train two fine models $f_\phi(X_\text{f}|\mathcal{I}_{X_\text{c}})$ and $f_{\phi'}(Y_\text{f}|\mathcal{I}_{Y_\text{c}})$ to fit the poses transformed into the corresponding ORoI space, $\psi_{X_\text{c}}(X)$ and $\psi_{Y_\text{c}}(Y)$, where $(X, Y)$ are sampled from the demonstrations. These fine models are conditioned on the corresponding ORoIs of pick and place targets, $\mathcal{I}_{X_\text{c}}$ and $\mathcal{I}_{Y_\text{c}}$, respectively. Similar to our coarse model, we represent the fine models using score networks $s_\phi(X_\text{f}|\mathcal{I}_{X_\text{c}}; \sigma)$ and $s_{\phi'}(Y_\text{f}|\mathcal{I}_{Y_\text{c}}; \sigma)$ respectively. Both networks operate on $SE(2)$.
\begin{figure*}[t]
  \centering
  \includegraphics[width=\textwidth]{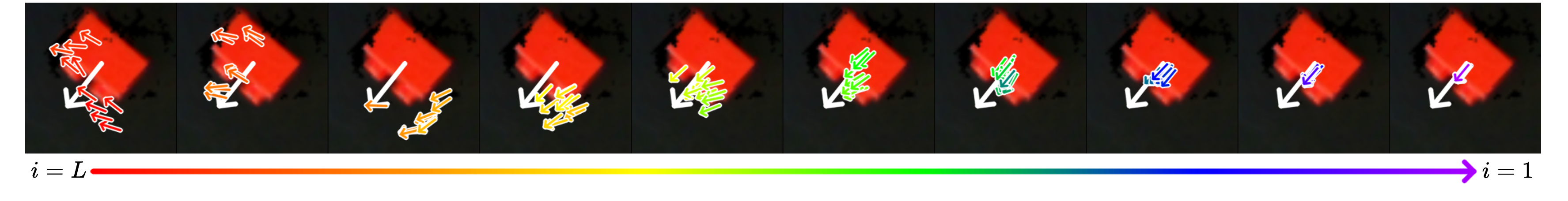}
  \caption{
  An illustration of the iterative refinement of pose estimates through denoising steps. White arrows represent the ground truth, while multi-colored arrows, transitioning from $i=L$ to $i=1$, signify the evolving pose estimate at each step.
  }
  \label{fig:diffusion-pose}
  \vspace{-1em}
\end{figure*}

We adopt a similar architecture as~\cite{hsiao2024confronting} and use the same design for our score networks, each of which comprises two main components: the conditioning part and the denoising part. In the conditioning part, the input image is encoded using ResNet~\cite{he2016deep} to generate a feature embedding. The time index $i$, representing the denoising time step, is encoded using positional embedding~\cite{vaswani2017attention}. These embeddings from the conditioning part are used to condition the neural networks in the denoising part. In the denoising part, the $N$ noisy poses in $SE(2)^N$ are transformed into Lie algebra representation and fed into a multilayer perceptron (MLP), which produces $N$ \textit{score} estimations. They are then used to calculate the DSM losses defined in Eq.~(\ref{eq:lie-dsm-n}) during the training phase and to denoise poses through the reverse process defined in Eq.~(\ref{eq:random-walk-n}) in the sampling phase. Fig.~\ref{fig:diffusion-pose} shows the pose denoising process. We specify $N=2$ for the coarse model and $N=1$ for the fine models.

\subsection{Data Augmentation}
To effectively train our diffusion model, it is essential to ensure that the training data distribution closely resembles the distribution encountered during inference. Nevertheless, due to the limited availability of demonstration data in various experimental scenarios, where the number of demonstrations can be as low as one, we incorporate augmentation techniques to expand the training dataset using only the available demonstration data for each case. We employ two distinct augmentation methods: (1) pose and (2) color augmentations.

Pose augmentation involves adjusting the target poses, which allows the model to learn from variations in object positions and orientations. This augmentation technique is implemented during the training of both the coarse and fine models. On the other hand, color augmentation is only applied to the fine model. It modifies the appearance of images by introducing variations in lighting, contrast, color balance, etc. This augmentation method is particularly beneficial for handling real-world scenarios characterized by inconsistent lighting conditions and camera noise. The integration of these augmentation strategies enhances our model's robustness and adaptability. As a result, our model can generalize effectively from limited data.  This results in improved performance and accuracy in real-world scenarios.
\begin{figure}[t]
  \centering
  \includegraphics[width=.9\linewidth]{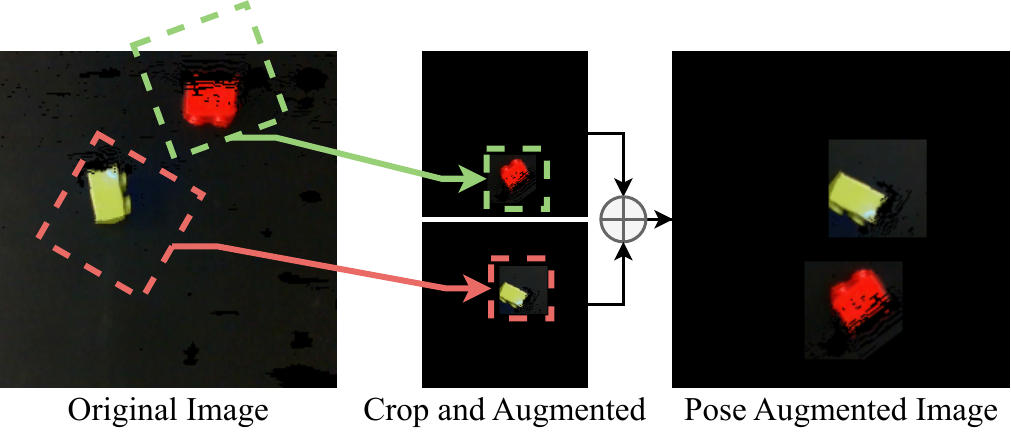}
  \caption{
  An illustration of the pose augmentation process that alters the pick and place poses for training the coarse model.
  }
  \label{fig:pose_aug}
  \vspace{-1em}
\end{figure}

More specifically, we implement two different pose augmentation approaches for the coarse model and the fine model. For the coarse model, we adopt a technique similar to~\cite{zeng2021transporter} and enhance its approach by applying augmentation to the pick and place poses separately instead of transforming both pick and place poses together using the same transformation. This is depicted in Fig.~\ref{fig:pose_aug}, in which we crop the pick-and-place objects from the images and apply random translations and rotations to emulate variations in their positions and orientations. This proposed pose augmentation method results in pick-and-place objects with various relative positions and angles. For the fine model, another pose augmentation strategy is designed to reflect the cropped and rotated images produced by the coarse pose estimates. Assuming that the coarse pose errors are relatively small, we crop and rotate the pick-and-place object regions according to the ground truth poses. To account for potential deviations, we further apply subtle random translations and rotations to these cropped regions. This augmentation approach emulates the variations resulting from the coarse pose estimates. This enables the fine model to learn and rectify minor pose errors during refinement.

\section{Experimental Results}
\label{sec:experiments}
\subsection{Environments}
\begin{figure}[t]
  \centering
  \includegraphics[width=.9\linewidth]{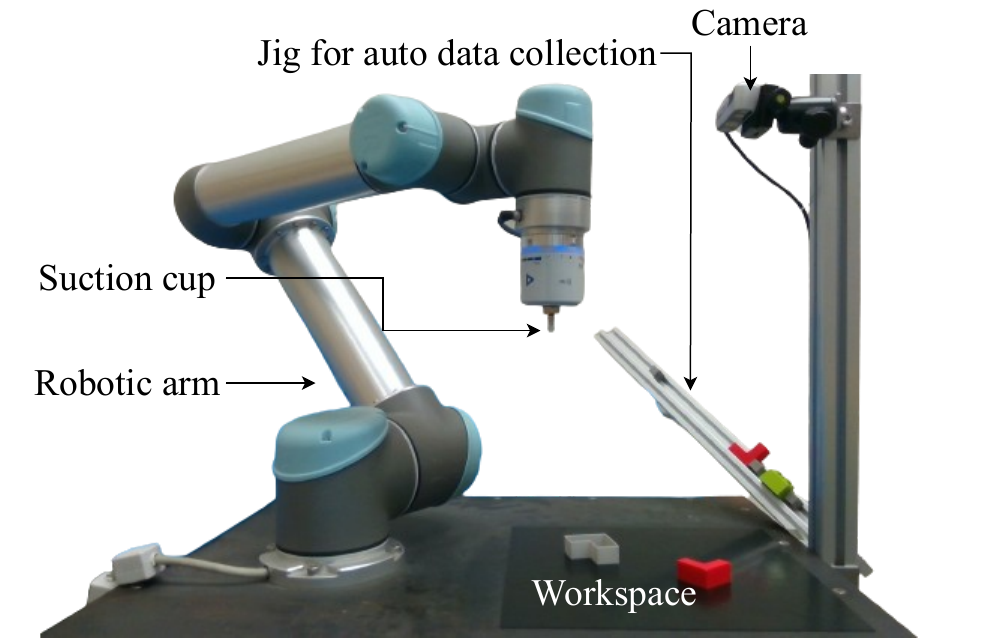}
  \caption{
  An illustration of the real robotic hardware setup.
  }
  \label{fig:robot}
  \vspace{-1em}
\end{figure}

\subsubsection{Environmental Setups}
We establish experimental environments in both simulation and the real world. In the simulation environments, we employ the Ravens simulator~\cite{zeng2021transporter}. The workspace size is set to $0.5 \times 1 \, \text{m}^2$, with three simulated RGB-D cameras directed towards the workspace. The simulated environments are adopted for collecting training and testing datasets. At test time, we directly evaluate the performance on the test dataset. For the real-world experiments, we arrange a workspace of size $0.224 \times 0.224 \, \text{m}^2$, and the entire robotic arm setup is illustrated in Fig.~\ref{fig:robot}, with an Intel RealSense depth camera D435i positioned above it. We utilize a Universal Robots UR5 robotic arm equipped with a Schmalz vacuum generator ECBPMi and an 8 mm diameter suction cup to perform our experiments. The use of the suction cup mitigates issues commonly encountered with grippers, such as inadvertently altering the object's position and orientation during gripping.

\begin{figure*}[t]
  \centering
  \includegraphics[width=\linewidth]{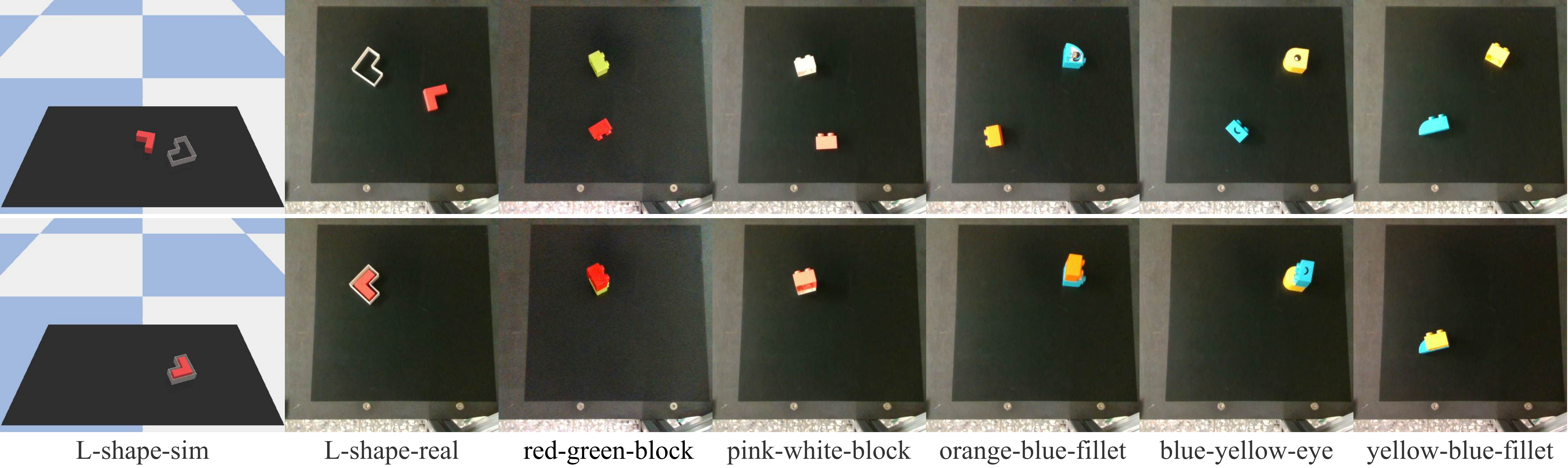}
  \caption{Simulation and real-world tasks. The top row depicts the initial states, while the bottom row shows the final states after task completion. The real-world tasks were executed using our methodology on a robotic arm.
  }
  \label{fig:task}
  \vspace{-1em}
\end{figure*}

\subsubsection{Tasks and Datasets}
We selected one task in the simulated environment and six tasks in real-world settings, as depicted in Fig.~\ref{fig:task}. For each task, we collected 100 training and 100 testing data samples. In the simulation environment, a block-insertion task named \textbf{L-shape-sim}, described in~\cite{zeng2021transporter}, was selected. This task aims to pick a red L-shape block and place it inside a gray L-shape frame. To collect the simulation dataset, we adapted the approach from~\cite{huang2024leveraging}, modifying discrete poses to continuous ones and picking at a fixed position on the L-shape block. This modification allows for a precise assessment of the proposed methodology's performance in terms of translational and rotational errors. The raw RGB-D images captured by three virtual cameras were top-down projected into an image of $320 \times 160$ pixels. In our study, we resized and padded images to $224 \times 224$ pixels.

In the real-world scenarios, we adapted the \textbf{L-shape-sim} task to a realistic robotic arm task named \textbf{L-shape-real}. Moreover, we conducted five additional challenging tasks that involve stacking LEGO DUPLO blocks on their sides: \textbf{(1) red-green-block}, stacking a red block on top of a green block; \textbf{(2) pink-white-block}, stacking a pink block on top of a white block; \textbf{(3) orange-blue-fillet}, stacking an orange block on top of a blue block with a fillet and printed drawing; \textbf{(4) blue-yellow-eye}, stacking a blue block with a closed-eye drawing on top of a yellow block with an eye drawing; and \textbf{(5) yellow-blue-fillet}, stacking a yellow block on top of a wider blue block with a fillet. For all real-world tasks except L-shape-real, we employed an automated data collection strategy that captures a set of pick-and-place images and actions approximately every 40 seconds. The robot picked up the objects from the jig, as shown in Fig.~\ref{fig:robot}, placed them in the workspace with random non-interfering pick-and-place poses, and returned them to their original positions on the jig after imaging. This process continued until data collection was complete. Due to the jig's tilted design, objects automatically realigned when placed back on the jig, which mitigated cumulative errors. For the L-shape-real task, we used a semi-automated data collection strategy that captures a set of pick-and-place images and actions approximately every 60 seconds. This strategy followed the same procedure as the automated strategy described above, except that the L-shape frame could not be lifted with a suction cup. As a result, the L-shape frame was manually aligned. Raw RGB-D images were top-down projected into $224 \times 224$ pixel images.

\subsection{Baselines}
We select Transporter Network~\cite{zeng2021transporter} and Equivariant Transporter~\cite{huang2024leveraging} as our baselines. Transporter Network is widely adopted as a baseline in related works, while Equivariant Transporter exhibits superior performance with fewer demonstrations. Both baselines are capable of accomplishing pick-and-place tasks with a limited number of demonstrations. Equivariant Transporter achieves higher sample efficiency due to its utilization of the equivariant network architecture. Although our method utilizes only RGB input, we trained and evaluated the baselines using top-down projected RGB-D images as per their original settings to minimize modifications, while our method uses top-down projected RGB images. The Transporter Network model is trained to 20,000 steps and evaluated at 20,000 steps, while the Equivariant Transporter model is trained to 10,000 steps and assessed at 10,000 steps.

\subsection{Training and Metrics}
\subsubsection{Training Procedure}
We train our coarse and fine models separately, using $K=\{1, 10, 100\}$ demonstrations sampled from the available data. The training and evaluation are performed on an Nvidia TITAN V GPU and an Intel Xeon E5-2620V4 CPU running at 2.10GHz. For the coarse model, the training takes approximately 40 minutes for 50,000 steps, while the training for the fine model requires around 50 minutes for 50,000 steps. Both models deploy a 34-layer ResNet~\cite{he2016deep} pre-trained on ImageNet~\cite{deng2009imagenet} for feature extraction from images. The training parameters used for both the coarse and fine models are as follows: the number of train steps is 50,000, with an initial learning rate of 1e-4 which is exponentially decayed to 1e-5, a batch size of 10, and 100 denoising steps. The inference times for the coarse and fine models are approximately 152ms and 165ms, respectively.

\subsubsection{Metrics}
\begin{figure}[t]
  \centering
  \includegraphics[width=.9\linewidth]{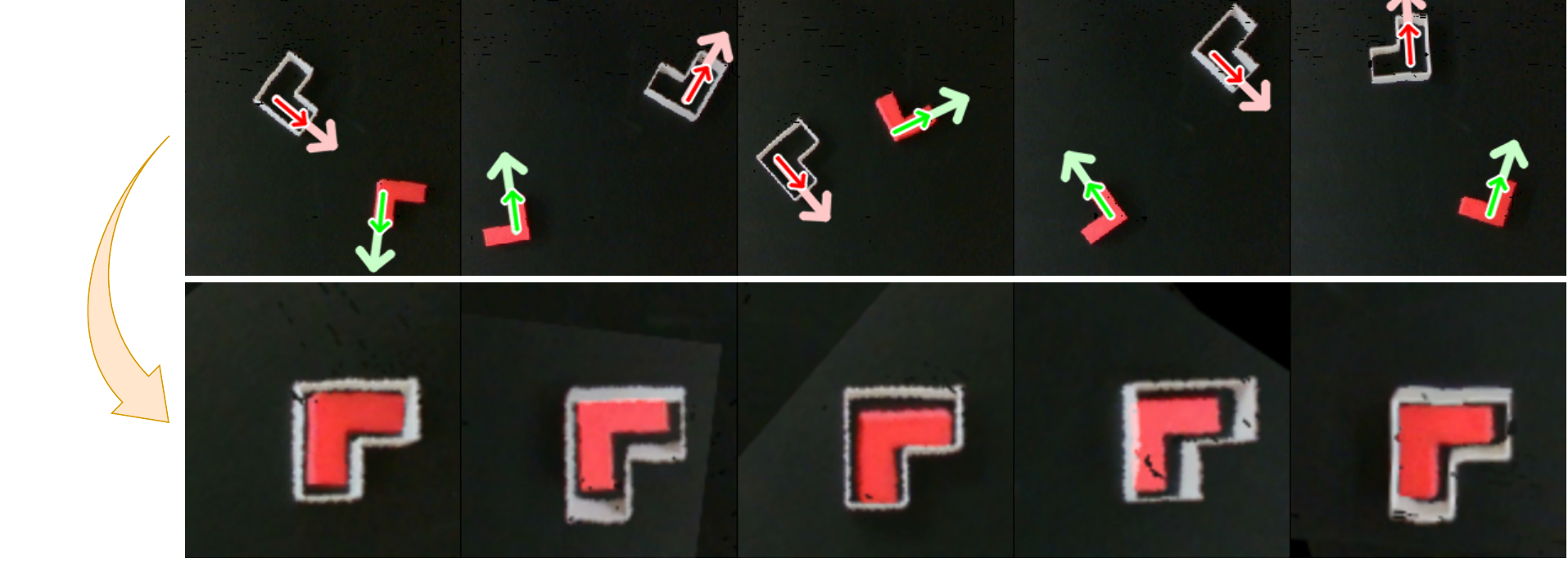}
  \caption{Simulated transport: The top row depicts the original image marked with green arrows for $X_\text{sum}$ and red arrows for $Y_\text{sum}$, with lighter-colored large arrows representing the ground truth. The bottom row shows the result after simulated transport by overlaying $X_\text{sum}$ and $Y_\text{sum}$ in the image space.}
  \label{fig:transport}
  \vspace{-1em}
\end{figure}
For the simulation experiment, we evaluate our methodology's performance on the 100 test data samples. We then calculate the pick pose error, the place pose error, and the transport pose error defined in Eq.~(\ref{eq:transport}) with respect to the ground truth. As shown in Fig.~\ref{fig:transport}, we visualize the before and after states of simulated transport in the image space. We consider a pick-and-place attempt successful if the pick translational error, place translational error, and transport translational error are all less than five pixels, and the transport rotational error is less than five degrees. The translational errors are computed in the image space coordinates of the projected camera views. For the real-world tasks, an error of one pixel corresponds to one mm in the physical workspace. On the other hand, for the simulation tasks, an error of one pixel corresponds to 3.125 mm in the simulated environment.

To validate that our method can achieve similar success rates in the real world as in the simulation, we deploy our method on a real robotic arm for pick-and-place evaluation. For each real-world task and for each model trained on $K=\{1, 10, 100\}$ demonstrations, we test ten randomly initialized scenarios. In each scenario, we place the pick and place objects in random poses within the workspace. The height for picking and placing is pre-set to an appropriate value for the objects being manipulated. The robotic arm then attempts to execute the pick-and-place transport by employing our method. The success of a pick-and-place task is determined based on the following criteria: for \textbf{L-shape-real}, whether the L-shape object is placed in the outer frame; for the LEGO blocks: (1) whether a block is stacked on another block without falling off, and (2) whether the final transport rotational error is less than five degrees.

\subsection{Performance Evaluation and Ablation Study}
\label{sec:results}
\subsubsection{Simulation}
\begin{table*}[t]
    \caption{Task success rate (\%) comparisons against baselines. The highest success rates are highlighted in bold.}
    \label{tables:success-rate}
    \centering
    \resizebox{\linewidth}{!}{%
        \begin{tabular}{l *{21}{wc{1.5em}}}
            \toprule
            \multicolumn{1}{r}{\textbf{Task}}                  & \multicolumn{3}{c}{\textbf{L-shape-sim}} & \multicolumn{3}{c}{\textbf{L-shape-real}} & \multicolumn{3}{c}{\textbf{red-green-block}} & \multicolumn{3}{c}{\textbf{pink-white-block}} & \multicolumn{3}{c}{\textbf{orange-blue-fillet}} & \multicolumn{3}{c}{\textbf{blue-yellow-eye}} & \multicolumn{3}{c}{\textbf{yellow-blue-fillet}}                                                                                                                                                                                                                                                                                                                                                                \\
            \cmidrule(lr){2-4} \cmidrule(lr){5-7} \cmidrule(lr){8-10} \cmidrule(lr){11-13} \cmidrule(lr){14-16} \cmidrule(lr){17-19} \cmidrule(lr){20-22}
            Method \hfill Demonstrations                       & \multicolumn{1}{c}{1}                    & \multicolumn{1}{c}{10}                    & \multicolumn{1}{c}{100}                      & \multicolumn{1}{c}{1}                         & \multicolumn{1}{c}{10}                         & \multicolumn{1}{c}{100}                      & \multicolumn{1}{c}{1}                          & \multicolumn{1}{c}{10} & \multicolumn{1}{c}{100} & \multicolumn{1}{c}{1} & \multicolumn{1}{c}{10} & \multicolumn{1}{c}{100} & \multicolumn{1}{c}{1} & \multicolumn{1}{c}{10} & \multicolumn{1}{c}{100} & \multicolumn{1}{c}{1} & \multicolumn{1}{c}{10} & \multicolumn{1}{c}{100} & \multicolumn{1}{c}{1} & \multicolumn{1}{c}{10} & \multicolumn{1}{c}{100} \\
            \midrule
            Transporter Network~\cite{zeng2021transporter}     & 25                                       & 67                                        & 87                                           & \textbf{16}                                   & 71                                             & 70                                           & 4                                              & 22                     & 70                      & 4                     & 8                      & 1                       & 0                     & 21                     & 62                      & 1                     & 18                     & 67                      & 0                     & 38                     & 69                      \\
            Equivariant Transporter~\cite{huang2024leveraging} & 62                                       & 91                                        & \textbf{93}                                  & 3                                             & 56                                             & 80                                           & 1                                              & 36                     & 64                      & 5                     & 24                     & 67                      & 1                     & 15                     & 76                      & 4                     & 20                     & 70                      & 0                     & 30                     & 62                      \\
            Coarse Stage (Ours)                                & 42                                       & 45                                        & 43                                           & 1                                             & 5                                              & 19                                           & 0                                              & 28                     & 42                      & 0                     & 30                     & 54                      & 0                     & 13                     & 41                      & 0                     & 42                     & 73                      & 1                     & 56                     & 62                      \\
            Coarse + Fine Stage (Ours)                         & \textbf{91}                              & \textbf{95}                               & \textbf{93}                                  & 11                                            & \textbf{93}                                    & \textbf{97}                                  & \textbf{19}                                    & \textbf{77}            & \textbf{99}             & \textbf{12}           & \textbf{77}            & \textbf{99}             & \textbf{8}            & \textbf{95}            & \textbf{98}             & \textbf{18}           & \textbf{66}            & \textbf{97}             & \textbf{6}            & \textbf{92}            & \textbf{99}             \\
            \bottomrule
        \end{tabular}
    }
\end{table*}
We compare the transport success rates of our method against the baselines in Table~\ref{tables:success-rate}. It is observed that our coarse + fine stage consistently achieves superior success rates across nearly all tasks and demonstration scenarios. Our method significantly outperforms the baselines. This is especially evident in scenarios involving ten demonstrations. In the case of \textbf{L-shape-sim}, where the simulation environment offers minimal noise and lighting interference, our method exhibits exceptional performance, even with only one demonstration. 
Despite the inferior success rate of the coarse stage, the high success rate achieved by the coarse + fine stage indicates that the ORoI crops $(\mathcal{I}_{X_\text{c}}, \mathcal{I}_{Y_\text{c}})$ derived from $X_\text{c}$ and $Y_\text{c}$ effectively serve as a successful initial guess. The performance of Transporter Network and Equivariant Transporter in \textbf{L-shape-sim} is inferior to that reported in the original papers. This discrepancy could be ascribed to several factors: (1) The original papers employed the best models in validation. (2) The success criteria in the original papers had a broader angle tolerance of fifteen degrees compared to five degrees in our setup. (3) In the Ravens simulator used in the original papers, up to three pick-and-place attempts were allowed while our experiments permit only a single attempt.
Moreover, while the two baselines show reasonable performance in simulated environments, they face challenges in real scenarios. The suboptimal performance of Transporter Network in the pink-white-block task could be attributed to the color similarity between the pink and white blocks, leading to confusion during picking and placing owing to their resemblance.
\begin{table*}[t]
    \caption{A comparison of the mean transport errors, all trained with ten demonstrations, with the lowest ones bolded.
    }
    \label{tables:mean-error}
    \centering
    \resizebox{\linewidth}{!}{%
        \begin{tabular}{l *{7}{cccccc}}
            \toprule
            \hfill\textbf{Task}                               & \multicolumn{2}{c}{\textbf{L-shape-sim}} & \multicolumn{2}{c}{\textbf{L-shape-real}} & \multicolumn{2}{c}{\textbf{red-green-block}} & \multicolumn{2}{c}{\textbf{pink-white-block}} & \multicolumn{2}{c}{\textbf{orange-blue-fillet}} & \multicolumn{2}{c}{\textbf{blue-yellow-eye}} & \multicolumn{2}{c}{\textbf{yellow-blue-fillet}}                                                                                                          \\
            \cmidrule(lr){2-3} \cmidrule(lr){4-5} \cmidrule(lr){6-7} \cmidrule(lr){8-9} \cmidrule(lr){10-11} \cmidrule(lr){12-13} \cmidrule(lr){14-15}
            Method \hfill Metric                              & pixel                                    & degree                                    & pixel                                        & degree                                        & pixel                                          & degree                                       & pixel                                          & degree       & pixel        & degree       & pixel        & degree       & pixel        & degree       \\
            \midrule
            Transporter Network\cite{zeng2021transporter}     & 1.1                                      & 4.1                                       & \textbf{2.7}                                 & 3.7                                           & 2.8                                            & 70.5                                         & 75.4                                           & 90.2         & 4.9          & 87.8         & 5.2          & 69.6         & \textbf{2.6} & 56.5         \\
            Equivariant Transporter\cite{huang2024leveraging} & \textbf{1.0}                             & \textbf{2.6}                              & 3.6                                          & 10.7                                          & \textbf{1.9}                                   & 63.5                                         & \textbf{2.2}                                   & 80.4         & 3.3          & 69.8         & 6.7          & 61.9         & 3.1          & 56.4         \\
            Coarse Stage (Ours)                               & 7.7                                      & 3.9                                       & 9.1                                          & 2.0                                           & 6.1                                            & 4.2                                          & 4.6                                            & \textbf{2.1} & 5.8          & 3.0          & 4.3          & 2.0          & 4.5          & 2.0          \\
            Coarse + Fine Stage (Ours)                        & 2.6                                      & 5.4                                       & \textbf{2.7}                                 & \textbf{1.0}                                  & 3.2                                            & \textbf{2.4}                                 & 2.7                                            & 7.7          & \textbf{2.4} & \textbf{1.6} & \textbf{3.7} & \textbf{1.8} & 2.7          & \textbf{1.4} \\
            \bottomrule
        \end{tabular}
    }
\vspace{-1em}
\end{table*}

In Table~\ref{tables:mean-error}, we present the mean transport translational and rotational errors for our methodology and the baselines, all trained on ten demonstrations. It is apparent that our coarse + fine stage achieves the lowest rotational error across most tasks, which substantiates the effectiveness of the fine stage in producing continuous and precise rotation refinement. Moreover, although the coarse stage's estimated translations and rotations exhibit slightly larger errors, the fine stage effectively corrects them. This may be credited to the fine stage's ability to filter out distractions and concentrate solely on the target object. Regarding translational errors, the baselines occasionally achieve slightly lower errors than our coarse + fine stage. However, their significantly higher rotational errors contribute to lower overall task success rates, as successful object transportation requires both precise translation and rotation predictions. We hypothesize that in these cases, the baselines obtain more accurate translations as the predicted place location is conditioned on the predicted pick location. This conditioning allows for the correction of inaccurate pick prediction when determining the place location. The results suggest that our coarse + fine stage shows excellent translational and rotational accuracy, which enables high pick-and-place success rates.

\subsubsection{Real Robot}
\begin{table}[t]
    \caption{The success rate (\%) of real-world tasks.}
    \label{tables:real-world}
    \newcolumntype{Y}{>{\centering\arraybackslash}X}
    \centering
    \resizebox{0.75\linewidth}{!}{%
        \begin{tabular}{l | *{3}{wc{3em}}}
            \toprule
                                       & \multicolumn{3}{c}{\textbf{Demonstrations}}                                                                      \\
            \textbf{Task}              & \multicolumn{1}{c}{\textbf{1}}              & \multicolumn{1}{c}{\textbf{10}} & \multicolumn{1}{c}{\textbf{100}} \\
            \midrule
            \textbf{L-shape-real}      & 40                                          & 90                              & 90                               \\
            \textbf{red-green-block}   & 40                                          & 100                             & 100                              \\
            \textbf{pink-white-block}  & 30                                          & 90                              & 90                               \\
            \textbf{orange-blue-fillet} & 50                                          & 100                             & 100                              \\
            \textbf{blue-yellow-eye}   & 60                                          & 100                             & 100                              \\
            \textbf{yellow-blue-fillet} & 40                                          & 100                             & 100                              \\
            \bottomrule
        \end{tabular}
    }
\vspace{-1em}
\end{table}

Table~\ref{tables:real-world} presents the pick-and-place success rates on our real robot for various numbers of demonstrations provided, tested with our coarse-to-fine methodology. The success rates align with the simulation results. The significantly higher success rates observed in comparison to simulation, particularly for scenarios involving one and ten demonstrations, may stem from the fact that successful stacking of LEGO blocks does not necessitate a precision of five mm, as set in the simulation criterion. Fig.~\ref{fig:task} illustrates several successful real-world pick-and-place instances executed by our methodology. These examples highlight our methodology's capability to achieve accurate object transportation. The observed success rates in real-world scenarios demonstrate the practical viability of our methodology for deployment in real robotic arm systems.

\subsubsection{Ablation Study}
\begin{figure}[t]
  \centering
  \includegraphics[width=.9\linewidth]{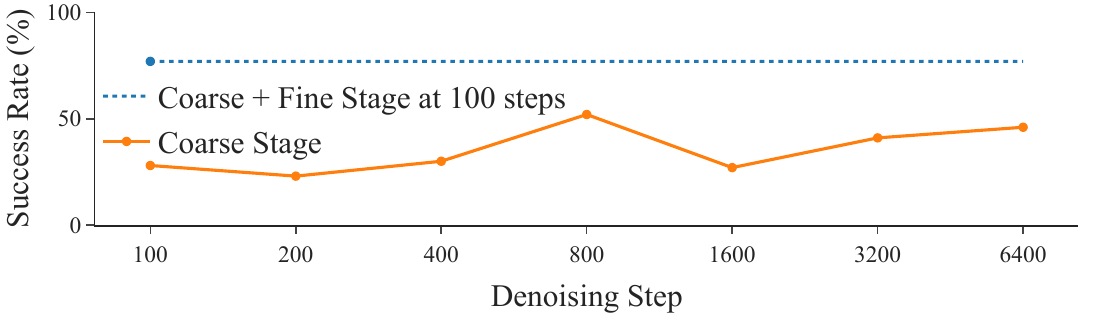}
  \caption{The success rate of the coarse stage vs. denoising step. Trained on the red-green-block task with ten demonstrations.}
  \label{fig:coarse-timestep}
  \vspace{-1em}
\end{figure}

We further provide experiments to evaluate the impact of different denoising steps on the coarse stage's performance. These ablation experiments focus on the red-green-block task and make use of ten demonstrations for training. As illustrated in Fig.~\ref{fig:coarse-timestep}, increasing the diffusion steps leads to a slight improvement in the transport success rate. Nevertheless, this improvement comes at the expense of increased inference time. Despite the improvement, the coarse + fine stage consistently outperforms the coarse stage with higher success rate. This observation highlights the fine stage's efficiency and efficacy in refining estimated poses.

\section{Conclusion}
\label{sec:conclusion}
In this work, we introduce a novel coarse-to-fine approach employing diffusion networks to augment the precision of pick-and-place operations in robotic manipulation tasks. Our methodology demonstrates exceptional performance in both simulated and real-world environments. It achieves high accuracy and success rates with minimal data requirements, relying solely on RGB-D top-down projected RGB images. We highlight the advantages of the coarse-to-fine strategy and analyze the distinct roles between the coarse and fine stages. Avenues for further exploration include the adoption of 3D pose or 2.5D pose estimation utilizing depth data, along with investigating non-top-down projected imagery.

\section*{Acknowledgments}
\label{sec:acknowledgments}
The authors gratefully acknowledge the support from the National Science and Technology Council (NSTC) in Taiwan under grant numbers MOST 111-2223-E-002-011-MY3, NSTC 113-2221-E-002-212-MY3, NSTC 113-2640-E-002-003, and NSTC 113-2922-I-007-247. The authors would also like to express their appreciation for the GPUs, donated by NVIDIA Corporation and NVIDIA AI Technology Center, used in this work. Furthermore, the authors extend their gratitude to the National Center for High-Performance Computing for providing the computational resources.

% \bibliographystyle{IEEEtranBST/IEEEtran} % use IEEEtran.bst style
% \bibliography{IEEEtranBST/IEEEabrv, reference}

\bibliographystyle{ieeetr}
\bibliography{reference}

\end{document}